\documentclass[10pt,twocolumn,letterpaper]{article}

\usepackage{cvpr}              

\usepackage{graphicx}
\usepackage{amsmath}
\usepackage{amssymb}
\usepackage{booktabs}

\usepackage[pagebackref,breaklinks,colorlinks]{hyperref}
\usepackage[linesnumbered, ruled]{algorithm2e}
\usepackage{multirow}
\usepackage{subfloat}

\usepackage[accsupp]{axessibility}  
\usepackage[capitalize]{cleveref}
\crefname{section}{Sec.}{Secs.}
\Crefname{section}{Section}{Sections}
\Crefname{table}{Table}{Tables}
\crefname{table}{Tab.}{Tabs.}

\newcommand*{\affmark}[1][*]{\textsuperscript{#1}}
\newcommand*{\email}[1]{\texttt{#1}}

\def\cvprPaperID{10192} 
\def\confName{CVPR}
\def\confYear{2022}

\begin{document}
\title{Graph-based Spatial Transformer with Memory Replay for Multi-future Pedestrian Trajectory Prediction}


\author{%
Lihuan Li\affmark[1] \quad Maurice Pagnucco\affmark[1] \quad Yang Song\affmark[1] \\
\affmark[1]University of New South Wales, Australia \\
\email{\tt\small lihuanli80@gmail.com \quad
 \{morri,yang.song1\}@unsw.edu.au}
}

\maketitle

\begin{abstract}
   Pedestrian trajectory prediction is an essential and challenging task for a variety of real-life applications such as autonomous driving and robotic motion planning. Besides generating a single future path, predicting multiple plausible future paths is becoming popular in some recent work on trajectory prediction. However, existing methods typically emphasize spatial interactions between pedestrians and surrounding areas but ignore the smoothness and temporal consistency of predictions. Our model aims to forecast multiple paths based on a historical trajectory by modeling multi-scale graph-based spatial transformers combined with a trajectory smoothing algorithm named ``Memory Replay'' utilizing a memory graph. Our method can comprehensively exploit the spatial information as well as correct the temporally inconsistent trajectories (e.g., sharp turns). We also propose a new evaluation metric named ``Percentage of Trajectory Usage'' to evaluate the comprehensiveness of diverse multi-future predictions. Our extensive experiments show that the proposed model achieves state-of-the-art performance on multi-future prediction and competitive results for single-future prediction. Code released at \href{https://github.com/Jacobieee/ST-MR}{https://github.com/Jacobieee/ST-MR}.
\end{abstract}
\section{Introduction}
Trajectory prediction is an indispensable part of social behavior analysis for a variety of applications including autonomous driving \cite{zhao2020tnt,chai2019multipath}, motion tracking \cite{sadeghian2017tracking,luo2018fast} and robotic systems \cite{rosmann2017online}. Such tasks require a high-level understanding of videos and human social behaviors to precisely forecast the future locations of pedestrians based on the observed trajectories and scenes. 

\begin{figure}[htb]
    \centering
    \subfloat[]{
    \label{intro}
    \includegraphics[width=3.5cm,height=2.5cm]{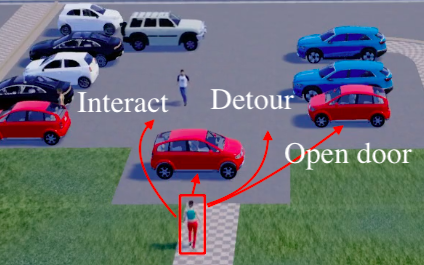}
    }
    \subfloat[]{
    \label{detour}
    \includegraphics[width=3.5cm,height=2.5cm]{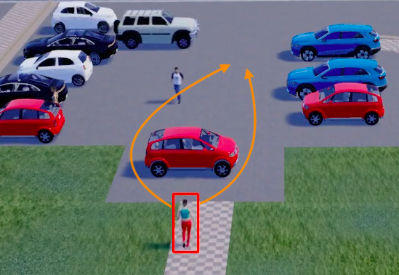}
    }
    \quad
    \subfloat[]{
    \label{obstacles}
    \includegraphics[width=3.5cm,height=2.5cm]{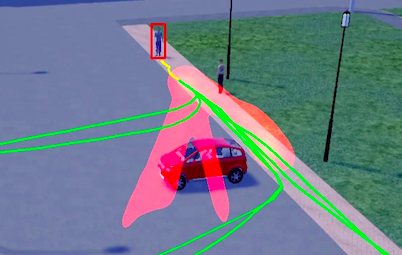}
    }
    \subfloat[]{
    \label{sharpturn}
    \includegraphics[width=3.5cm,height=2.5cm]{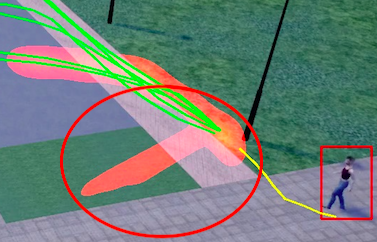}
    }
    
    \setlength{\abovecaptionskip}{0.cm}
    \setlength{\belowcaptionskip}{-0.7 cm}
    \caption{Illustration of multi-future trajectory prediction and existing issues. The yellow and green lines are the observed and ground truth trajectories. \subref{intro} Multiple future trajectories (red arrows) influenced by different intentions and destinations. \subref{detour} Multiple options of paths (orange arrows) heading to the same destination. \subref{obstacles} An imperfect prediction (with heatmap) in Multiverse \cite{liang2020garden} that goes through a vehicle. \subref{sharpturn} An imperfect prediction (with heatmap in the red circle) in Multiverse \cite{liang2020garden} that violates temporal consistency.}
    \label{fig:intro_figures}
\end{figure}

Trajectory prediction requires simultaneous processing of spatial and temporal information. While walking paths naturally exhibit a temporal consistency, it is also important to model spatial interactions among pedestrians such as talking, grouping and avoiding collisions. Other objects in a scene also affect the future paths, as pedestrians tend to avoid obstacles (e.g., street lamps, trees, vehicles) or unnecessary change of paths (e.g., walk from the pavement to the middle of the road). However, reactions to spatial interactions may also undermine the original intentions based on temporal information. Even if both of them are properly processed, it is still a conundrum to predict spatially reasonable trajectories while conforming to temporal consistency.

Real-world datasets \cite{awad2018trecvid,pellegrini2010improving,lerner2007crowds} have enabled the research on trajectory prediction and current approaches \cite{liang2019peeking,gupta2018social} have made great progress on single-future trajectory prediction, where the predicted trajectories are evaluated against the ground truth trajectories recorded in the videos. However, the human mind is capricious and realistic situations are complicated. Given an observed trajectory, there can be multiple different destinations and multiple plausible future trajectories. Fig. \ref{fig:intro_figures}\subref{intro} demonstrates that multiple intentions and destinations can drive the pedestrian at the bottom to walk in different paths. Fig. \ref{fig:intro_figures}\subref{detour} shows the pedestrian can select different paths to the same destination. 

To evaluate models that generate multiple possible trajectories, Liang \etal \cite{liang2020garden} recently proposed a simulated dataset named ``Forking Paths'', which provides multiple ground truth trajectories for the same historical trajectory. And in the same work, a two-stage end-to-end probabilistic model named ``Multiverse'' is designed for multi-future trajectory prediction. However, there are still some issues with this model. For example, Fig. \ref{fig:intro_figures}\subref{obstacles} shows a path through the vehicle; Fig. \ref{fig:intro_figures}\subref{sharpturn} is an example of temporal inconsistent prediction which violates the normal pattern of human motion with a sharp turn.

In this paper, we propose an encoder-decoder network to address the aforementioned issues. To effectively process the spatial information, we first construct a multi-scale graph to represent scene segmentation and trajectory features. Then, we design a graph-based spatial transformer to learn the interactions between a pedestrian and other pedestrians as well as scene objects. Moreover, to integrate global temporal information, we develop a ``Memory Replay'' algorithm, which utilizes a memory graph to accumulate temporal information and ``replay'' it to the transformer at each time step to ensure the smoothness of trajectories. In addition, we propose a new evaluation metric ``Percentage of Trajectory Usage'' to evaluate the comprehensiveness of multi-future prediction, to complement the existing minADE$_{K}$ and minFDE$_{K}$ metrics in \cite{liang2020garden}. We show that our model achieves state-of-the-art performance on multi-future prediction on the Forking Paths dataset; and our results on single-future prediction are comparable to the current state-of-the-art models on the VIRAT/ActEV\cite{awad2018trecvid} dataset. We summarize our main contributions as follows:
\begin{enumerate}
    \vspace{-1.ex}
    \item We propose a graph-based spatial transformer for spatial interactions of pedestrians. By integrating the attention mechanism and graph structure, the spatial transformer can comprehensively generate and aggregate spatial features.
    \vspace{-1.ex}
    \item We design a novel trajectory smoothing algorithm, Memory Replay, for improving the temporal consistency of predicted trajectories and minimizing the conflicts between spatial and temporal information.
    \vspace{-1.ex}
    \item We define a new evaluation metric, Percentage of Trajectory Usage (PTU), to evaluate the comprehensiveness of multi-future prediction.
\end{enumerate}

\section{Related Work}
\noindent \textbf{Pedestrian trajectory prediction.}
There have been various methods that aim to forecast multiple possible future trajectories. Recent approaches \cite{gupta2018social,sadeghian2019sophie,kosaraju2019social} apply Generative Adversarial Networks (GANs) to generate a distribution of trajectories. Inverse Reinforcement Learning (IRL) \cite{kitani2012activity,liu2021social,choi2020regularising} is also becoming popular on multi-future trajectory prediction tasks. Besides, predicting multiple trajectories is emerging in vehicle trajectory prediction \cite{tang2019multiple,chai2019multipath,li2019grip++,zhao2020tnt}. These approaches, however, have been evaluated using single-future trajectories, as the ground truth contains a single path for each pedestrian. Currently, the \textit{Multiverse} model \cite{liang2020garden} achieves the state-of-the-art performance on the new 3D simulated dataset, the \textit{Forking Paths}, which is the first public benchmark designed specifically for evaluating the generation of multi-future trajectories. Our model outperforms \textit{Multiverse} on the \textit{Forking Paths} dataset for multi-future trajectory prediction.

\noindent \textbf{GNN-based models.}
In recent years, Graph Neural Networks (GNNs) have become popular. Traditional GNN models such as Graph Convolutional Network (GCN)\cite{kipf2016semi}, GraphSAGE\cite{hamilton2017inductive} and Graph Attention Network (GAT)\cite{velivckovic2017graph} are widely used in computer vision tasks such as pose estimation \cite{yan2018spatial,zhao2019semantic}, panoptic segmentation \cite{wu2020bidirectional}, point cloud analysis \cite{zhao2019pointweb}, etc. For pedestrian trajectory prediction, Sun \etal \cite{sun2020recursive} construct a GCN-based recursive social behavior graph (RSBG) given the annotations by sociologists. STGAT \cite{huang2019stgat} models a spatial-temporal graph attention network to encode the pedestrian interaction. Other works \cite{ivanovic2019trajectron,mohamed2020social,zhang2019sr,haddad2020graph2kernel} also implement improvements on GNNs to contribute to pedestrian trajectory prediction. We construct a multi-scale graph to model the interactions between pedestrians and multiple scales of surrounding areas.

\noindent \textbf{Transformer-based methods.}
Transformer-based methods \cite{vaswani2017attention} have been a trend in deep learning tasks. It was first used in Natural Language Processing \cite{devlin2018bert,wang2018glue,radford2018improving}, then flourishes in computer vision \cite{fu2019dual,dosovitskiy2020image,carion2020end,zheng2021rethinking}. Modeling both spatial and temporal transformer \cite{yu2020spatio,zhao2020spatial} can compete or even outperform the traditional sequence-to-sequence models in trajectory prediction, demonstrating their effectiveness in complex spatio-temporal feature processing. Other methods \cite{yuan2021agentformer,li2020social,bertugli2021ac} have also inserted transformer-based modules and achieved high performance on both pedestrian and vehicle trajectory prediction. We design a novel graph-based spatial transformer containing an attention-based message generation and a GAT-based aggregation method to effectively collect and process spatial information.


\section{Methods}
\begin{figure*}[htb]
\vspace{-0.5 em}
\centering

\includegraphics[scale=0.46]{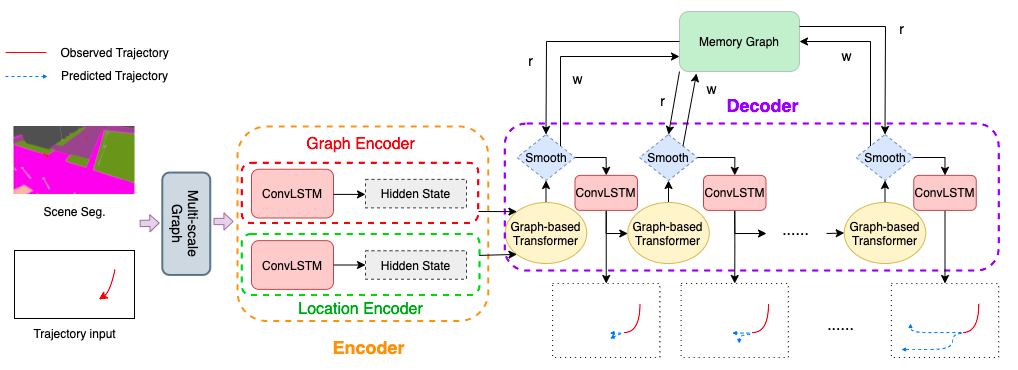}
\setlength{\abovecaptionskip}{-0.2 cm}
\setlength{\belowcaptionskip}{-0.5 cm}
\caption{Overview of our model. Graph encoder and location encoder each encodes the node level and coordinate level features processed by the multi-scale graph. At each time step in decoding, our proposed graph-based spatial transformers infer the possible neighboring locations at the next step. Then, our trajectory smoothing algorithm smooths and corrects the locations that violate the temporal consistency based on a memory graph that stores the temporal information of the earlier trajectory.}

\label{fig:overview}

\end{figure*}
\subsection{Overview}
Given a series of scene semantic segmentation maps $S=S_{1},S_{2},\ldots,S_{T_{obs}}$ and positions $X=(x_{1},y_{1}),(x_{2},y_{2}),\ldots,(x_{T_{obs}},y_{T_{obs}})$ of a pedestrian for time $1:T_{obs}$, our model aims to predict multiple possible future trajectories where the $i$th prediction of a pedestrian is denoted as $\hat{Y^{i}}=(\hat{x}^i_{t},\hat{y}^i_{t})$ for time $t=T_{obs+1}:T_{pred}$ by learning and inferring $P(\hat{Y}|S,X)$.

Fig.~\ref{fig:overview} shows the overall structure of our proposed model. Our model takes the observed trajectory $X$ and scene segmentation $S$ for video frames in the observation period as inputs. A multi-scale graph is constructed, which is a two-dimensional grid where the areas of the grid cells change with different scales. Each grid cell contains a sub-area of the scene segmentation and trajectory information. The inputs processed by the multi-scale graph are passed into an encoder-decoder network to generate future trajectories. The encoder encodes the motion pattern of pedestrians as well as the scene feature over time. The decoder consists of two main components: a multi-scale graph-based spatial transformer and a trajectory smoothing algorithm which is named ``Memory Replay''. The spatial transformer processes the information and makes prediction of the next step. Memory Replay smooths the prediction by reading and writing in a memory graph that contains the overall temporal information of the decoded trajectory. At each time step, the decoder generates a probability distribution of locations at the next time step using these two components, followed by a convolutional LSTM cell \cite{xingjian2015convolutional}. The most probable locations that we select at each time is determined by the diverse beam search \cite{li2016simple}.

\subsection{Multi-scale Graph Generation}We formulate a video frame as a graph $ \mathcal{G}(V,E)$ with a set of nodes $V$ and edges $E$. Specifically, we use a 2D regular grid with $|V|$ grid cells to split a frame into multiple areas, where each area can be considered a node $v \in V$ that connects to adjacent nodes with an undirected edge $e \in E$. Each grid cell can establish connections to the horizontal, vertical and diagonal neighbors. Inspired by the idea of feature pyramid \cite{lin2017feature,liang2020garden}, we design the graph in different scales to process spatial information in multiple levels. There are two scales of grids which are the same as those in \cite{liang2020garden}, and subsequently, the number of nodes can be $36\times18$ and $18\times9$.

Our implementation differs from feature pyramid where we change the amount of features included in a node instead of resizing the whole image (video frame). Nodes in the larger scale have less but finer features, and nodes in the smaller scale have more but coarser features. By learning on the multi-scale graph, our model can be more adaptive to different levels of information and make comprehensive decisions based on surrounding areas of a pedestrian.


\subsection{Spatio-temporal Encoder}
Inspired by recent studies \cite{ren2016faster,liang2020garden}, we propose two types of trajectory encoder: graph encoder and location encoder. In each graph scale, the graph encoder encodes the node level feature which is the index of the grid cell where the current location belongs to, and the location encoder record the specific coordinate which is the offset from the center of the area covered by the node. These two encoded hidden states are passed into the decoder separately.

In contrast to recent approaches \cite{gupta2018social,sun2020recursive,alahi2016social} that model the motion of all pedestrians in a scene to enrich the spatial features, we utilize convolutional LSTM \cite{xingjian2015convolutional} to encode both spatial and temporal features simultaneously:
\begin{align}
    \setlength{\abovedisplayskip}{3pt}
    \setlength{\belowdisplayskip}{3pt}
    H(g)^\mathcal{G}_{t}&={\rm ConvLSTM}(g^\mathcal{G}_{t}, H(g)^\mathcal{G}_{t-1}) \label{g_encoder}
    \\
    H(l)^\mathcal{G}_{t}&={\rm ConvLSTM}(l^\mathcal{G}_{t}, H(l)^\mathcal{G}_{t-1}) \label{l_encoder}
\end{align}
which denote the hidden states of graph encoder and location encoder at the graph scale $\mathcal{G}$ at time $t$, respectively. Since the subsequent process in the encoder and decoder all operates in the same way on these two hidden states, we denote both hidden states collectively as $H^\mathcal{G}_{t}$. To embed the pedestrian location $(x_{t},y_{t})$ for the graph encoder in graph scale $\mathcal{G}$ at time $t$, we adapt one-hot encoding multiplied by the scene segmentation:
\begin{equation}
    \setlength{\abovedisplayskip}{3pt}
    \setlength{\belowdisplayskip}{3pt}
    g^\mathcal{G}_{t}=\text{one-hot}(\text{idx}(x_{t},y_{t})^\mathcal{G}) \odot S^\mathcal{G}_{t} 
\end{equation}
where the idx$()$ function converts the coordinate to the index of the grid cell in $\mathcal{G}$. Then, the one-hot function projects the indexed cell onto the corresponding position on the graph. For the location encoder, we further calculate the offset from the center of the indexed cell:
\begin{equation}
    \setlength{\abovedisplayskip}{3pt}
    \setlength{\belowdisplayskip}{3pt}
    l^\mathcal{G}_{t}=x^{\prime^{\mathcal{G}}}_{t},y^{\prime^{\mathcal{G}}}_{t}=(x_{t},y_{t})-C(\text{idx}(x_{t},y_{t})^\mathcal{G})
\end{equation}
where $C()$ function obtains the center coordinate of the indexed cell. The offset $(x^{\prime^{\mathcal{G}}}_{t},y^{\prime^{\mathcal{G}}}_{t})$ can be calculated by subtracting such center coordinate from the real coordinate $(x_{t},y_{t})$. We denote this offset coordinate as $l^\mathcal{G}_{t}$.

We keep the hidden state $H^\mathcal{G}_{T_{obs}}$ at time $T_{obs}$ after finishing encoding. And following \cite{liang2020garden}, we also calculate an average of semantic segmentation map $\Bar{S^\mathcal{G}}=\frac{1}{T_{obs}}\sum^{T_{obs}}_{t=1} S^\mathcal{G}_{t}$ and construct the hidden state to be passed into the decoder under graph scale $\mathcal{G}$ as:
\begin{equation}
    \setlength{\abovedisplayskip}{3pt}
    \setlength{\belowdisplayskip}{3pt}
    H^\mathcal{G}_{T_{obs}}=(H^\mathcal{G}_{T_{obs}} || \Bar{S^\mathcal{G}})
    \label{final_hidden}
\end{equation}
where $||$ is concatenation. The scene segmentation map provides the pedestrians with awareness of the content and location for each object in the scene, which facilitates modeling of human-to-scene interactions.

\subsection{Graph-based Spatial Transformer}
Although RNN models are commonly used for addressing sequence prediction tasks \cite{huang2019stgat}, they have limitations in collecting nearby information of a person. Recently, \cite{li2020social,zhao2020spatial} have made some progress on trajectory prediction by exploiting attention mechanism spatially and temporally. \cite{huang2019stgat} has employed GAT to model human-to-human relationship. However, compared to the crowded scenes in the ETH \cite{pellegrini2010improving} and UCY \cite{lerner2007crowds} datasets, most scenes in the VIRAT/ActEV and Forking Paths datasets \cite{awad2018trecvid,liang2020garden} have much fewer pedestrians, whereas the spatial interaction with scene objects is also important. 

Therefore, we design a graph-based spatial transformer to effectively model the relationship between both human-to-scene and human-to-human with the help of scene semantic segmentation features. The transformer takes the graph structured encoded hidden states $H^\mathcal{G}_{T_{obs}}$ from Eq. \eqref{final_hidden} as input node states. We use an attention mechanism to generate messages for all node pairs and aggregate them by the graph structure. The transformer will finally produce a group of updated node states that indicates the possible locations for the next time step.

\noindent \textbf{Attention-based message generation.} We generate two types of messages from node $v_{j}$ to node $v_{i}$: attention message and global message.

To extract the interactions between the pedestrian and neighboring areas, we first generate an attention message:

\begin{equation}
    \setlength{\abovedisplayskip}{3pt}
    \setlength{\belowdisplayskip}{3pt}
    M^{\mathcal{G}}_{Attn[i \leftarrow j]}=f^\mathcal{G}_{V[i]} \odot (f^\mathcal{G}_{Q[i]} || f^{\mathcal{G}^T}_{K[j]})+\Vec{b}^\mathcal{G}
    \label{msg}
\end{equation}
where $M^\mathcal{G}_{Attn}$ is a matrix containing messages between all pairs of nodes in the graph scale $\mathcal{G}$. We learn a query matrix $f^\mathcal{G}_{Q}$, key matrix $f^\mathcal{G}_{K}$ and value matrix $f^\mathcal{G}_{V}$ from the graph structured hidden state $H^\mathcal{G}_{T_{obs}}$. Inspired by GAT, we concatenate the query matrix and the transpose of key matrix to create an entry for calculating attention value of each node pair $v_{i}$ and $v_{j}$. Then, we assign an importance value to each entry by conducting an element-wise multiplication between the concatenated matrix and the value matrix and add bias $\Vec{b^\mathcal{G}}$. We denote the message from node $v_{j}$ to node $v_{i}$ as $M^\mathcal{G}_{Attn[i \leftarrow j]}$.

The spatial transformer adapts both advantages of self-attention mechanism and graph structure so that a pedestrian can attach different importance on the neighboring areas. However, distant objects (e.g., people, vehicles, obstacles) also provide essential spatial context in trajectory planning. Therefore, we also calculate a similarity score between each node pair $v^\mathcal{G}_i$ and $v^\mathcal{G}_j$ as a global message:
\begin{equation}
    \setlength{\abovedisplayskip}{3pt}
    \setlength{\belowdisplayskip}{3pt}
    M^\mathcal{G}_{global[i \leftarrow j]}=\Vec{h}^\mathcal{G}_{i}\Vec{h}^{\mathcal{G}^T}_{j} \label{simi_score}
\end{equation}
where $M^\mathcal{G}_{global[i\leftarrow j]}$ estimates the distance between features of node $v_i$ and $v_j$ in the hidden space. Finally, we obtain the total message passed from node $v_{j}$ to node $v_{i}$ by summing these two types of messages together:
\begin{equation}
    \setlength{\abovedisplayskip}{3pt}
    \setlength{\belowdisplayskip}{3pt}
    M^\mathcal{G}_{[i \leftarrow j]}=M^\mathcal{G}_{Attn[i \leftarrow j]} \oplus M^\mathcal{G}_{global[i\leftarrow j]} 
    \label{total_msg}
\end{equation}
where $\oplus$ is element-wise addition. The total message includes both unique information from neighboring nodes and similarity estimation from a global view.

\noindent \textbf{Update of node states.}
To update node states, we first calculate the edge weights based on the total message in Eq. \eqref{total_msg}:
\begin{equation}
    \setlength{\abovedisplayskip}{3pt}
    \setlength{\belowdisplayskip}{3pt}
    e^\mathcal{G}_{[i \leftarrow j]}=\frac{\exp(M^\mathcal{G}_{[i \leftarrow j]})}
    {\sum_{k \in \mathcal{N}^\mathcal{G}_{i}} \exp(M^\mathcal{G}_{[i \leftarrow k]})
    }
    \label{attn_coef}
\end{equation}
where calculated edge weight $e^\mathcal{G}_{[i \leftarrow j]}$ is normalized by the softmax function and node $k$ belongs to the neighbors of node $i$. To update the new node states, we apply a simple dot product between calculated edge weight and the node state in the previous time step:
\begin{equation}
    \setlength{\abovedisplayskip}{3pt}
    \setlength{\belowdisplayskip}{3pt}
    \widetilde{H}^\mathcal{G}_{t}(i)={e^\mathcal{G}_{[i \leftarrow j]}}h^\mathcal{G}_{i} \label{gat}
\end{equation}
where $\widetilde{H}^\mathcal{G}_{t}$ is the calculated node state for all nodes. The output at time $t$ and the new hidden state at time $t+1$ are generated by:
\begin{align}
    \setlength{\abovedisplayskip}{3pt}
    \setlength{\belowdisplayskip}{3pt}
    \hat{P}^\mathcal{G}_{t}&=\sigma(\delta{1}(H^\mathcal{G}_{t})) \label{node_states} \\
    H^\mathcal{G}_{t+1}&={\rm ConvLSTM}(\widetilde{H}^\mathcal{G}_{t},\delta_{2}(\hat{P}^\mathcal{G}_{t})) \label{new_hidden}
\end{align}
where $\hat{P}^\mathcal{G}_{t}$ is the prediction at time $t$. For each node $v_{i}$, $\hat{P}^\mathcal{G}_{t}(i)$ can be considered a probability if the input is from the graph encoder, or a coordinate value offset from the center of node $v_{i}$ if the input is from the location encoder. The output at time $t$ will be the input at $t+1$ and be passed into the convolutional LSTM cell with the updated node states $\widetilde{H}^\mathcal{G}_{t}$. $\delta_{1}$ and $\delta{2}$ are two different linear layers. The hidden state at time $t+1$ is represented as $H^\mathcal{G}_{t+1}$.



\subsection{Memory Replay} \label{smooth}
Our spatial transformer can encourage the model to focus more on the most probable areas but neglects to take into account the temporal consistency. In the decoder, decoded hidden states from the transformer at time step $t$ is heavily based on the decoded states at $t-1$. However, the locations at the current time is also influenced by the hidden states at all previous time steps. In other words, if we only consider the computations based on the most recent time step, our predictions sometimes deviate from the originally intended destinations implied by the hidden states further before. To address this issue, we propose a trajectory smoothing algorithm ``Memory Replay'' that utilizes a memory graph $G(V)$ to dynamically record the decoded temporal information of a trajectory, where $|V|$ is the same as the number of nodes in the hidden state graph scale $\mathcal{G}$. Memory Replay operates on the edge weights calculated by the transformer. At each time step, the memory graph $G$ carries the smoothed edge weights of all node pairs (including a node to itself) in the past decoding time steps and decreases the weights of edges that point to the temporally inconsistent locations.

The processing steps are shown in Algorithm \ref{memreplay}. Before decoding, we initialize the memory graph to all zeros. During each time step in decoding, we first pass the hidden state at the previous time into our spatial transformer to calculate the edge weights $e^\mathcal{G}$ for each node by Eq. \eqref{attn_coef} (Line 5). Then, we smooth the value of $e^\mathcal{G}$ by element-wise addition with $G$ (Line 6). $G$ is populated with the smoothed $e^\mathcal{G}$ at every time step (Line 7) to ensure that it includes the most recent decoded states. The hidden state at the current time is calculated by combining the updated node states based on the smoothed edge weights and the hidden state at the previous time in Eq. \eqref{gat}, as well as the output at the previous time (observed trajectory when the time is $T_{obs}$) in  Eq. \eqref{new_hidden} (Line 9). Finally, we generate the output at the current time based on the new hidden state by Eq. \eqref{node_states} (Line 10). Therefore, the memory graph can record the newest edge weights at each time step, where the edge weights are smoothed by the memory graph at the previous time step. Memory Replay effects in such a recursive manner.


\IncMargin{0em}
\begin{algorithm}
    \DontPrintSemicolon
    \caption{Memory Replay.}
    \label{memreplay}
    \SetKwInOut{Input}{input}\SetKwInOut{Output}{output}

    \Input{Encoded last hidden state $H^\mathcal{G}_{T_{obs}}$ and the last observed trajectory $\hat{P}^\mathcal{G}_{T_{obs}}$ in Graph scale $\mathcal{G}$ at time $T_{obs}$}
    \BlankLine
    \For{$\mathcal{G} \in $\{$(18, 32), (9, 16)$\}}{
        $G \leftarrow zeros \times \mathcal{G}$\;
        $H^\mathcal{G}_{T_{prev}}, \hat{P}^\mathcal{G}_{T_{prev}} \leftarrow H^\mathcal{G}_{T_{obs}}, \hat{P}^\mathcal{G}_{T_{obs}}$\;
        \For{$T_{curr} \longleftarrow T_{obs+1}$ \KwTo $T_{pred}$}{
            $e^\mathcal{G} \leftarrow $ calculate edge weights by the spatial transformer in Eq. \eqref{attn_coef}\;
            $e^\mathcal{G} \leftarrow \sigma(e^\mathcal{G} \oplus  G)$\;
            $G \leftarrow e^\mathcal{G}$\;
            
            $T_{prev} \leftarrow T_{curr}$\;
            $H^\mathcal{G}_{T_{prev}} \leftarrow$ prepare for hidden state in the next time step by Eq. \eqref{gat} and Eq. \eqref{new_hidden} \;
            $\hat{P}^\mathcal{G}_{T_{prev}} \leftarrow$ generate output with $H^\mathcal{G}_{T_{prev}}$ by Eq. \eqref{node_states}
        }
    }

\end{algorithm}
\DecMargin{0 em}
\setlength{\textfloatsep}{0.1cm}
\setlength{\floatsep}{-0.5 cm}
\subsection{Loss}
Following \cite{liang2020garden}, we split our training as a classification task (graph encoder stream) and regression task (location encoder stream). We consider the ground truth output for each graph scale $\mathcal{G}$ at each time $t$ as $P^\mathcal{G}_{i}(t)$ and the duration for loss calculation as $T_{1:loss}$. We use cross-entropy loss for the graph encoder stream:
\begin{equation}
    \setlength{\abovedisplayskip}{3pt}
    \setlength{\belowdisplayskip}{3pt}
    L^\mathcal{G}_{c}= -\frac{1}{T_{loss}} \sum^{T_{loss}}_{t=T_{1}} \sum_{i \in \mathcal{G}} P^\mathcal{G}_{i}(t) \log(\hat{P}^\mathcal{G}_{i}(t))
\end{equation}
In addition, inspired by \cite{sun2020recursive}, we propose exponential smooth $L1$ loss for the location encoder stream:
\begin{equation}
    \setlength{\abovedisplayskip}{3pt}
    \setlength{\belowdisplayskip}{3pt}
    L^\mathcal{G}_{r}= \frac{1}{T_{loss}} \sum^{T_{loss}}_{t=T_{1}} \sum_{i \in \mathcal{G}} {\rm Smooth}_{L1}(P^\mathcal{G}_{i}(t),\hat{P}^\mathcal{G}_{i}(t))\times e^{\frac{T_{loss}-t+1}{\mu}}
\end{equation}
where we define a penalty term $e^{\frac{T_{loss}-t+1}{\mu}}$ to guide the model focus more on the predictions at earlier time steps as the quality of earlier trajectories can greatly affect the later ones. The hyperparameter $\mu$ is used to control the strength of the penalty term.

To benefit from the multi-scale graph, we refer to multi-scale discriminators \cite{wang2018high} and sum the loss calculated in both scales \textit{Scales} $\in [36 \times 18, 18 \times 9]$:
\begin{equation}
    \setlength{\abovedisplayskip}{3pt}
    \setlength{\belowdisplayskip}{3pt}
    L=\sum_{\mathcal{G} \in Scales} \alpha L^\mathcal{G}_{c}+\beta L^\mathcal{G}_{r}
\end{equation}
and $L$ is used to optimize the training in both scales. 

\subsection{Generation of Multiple Trajectories}
We refer to \cite{li2016simple,liang2020garden} utilizing diverse beam search to generate multiple trajectories in the graph encoder stream. At time $t-1$ in the graph scale $\mathcal{G}$, we obtain a set of $K$ decoded trajectories with their conditional logarithmic probabilities denoted as ${C^{\mathcal{G},k}_{1},C^{\mathcal{G},k}_{2}\ldots C^{\mathcal{G},k}_{t-1}}$ where $k \in [1,K]$ and $K$ is the beam size. Given the probability $\hat{P}^{\mathcal{G},k}_{t}$ inferred by the model at time $t$, we calculate the new logarithmic probability of the graph node $i$ in the beam $k$ as:
\begin{equation}
    \setlength{\abovedisplayskip}{3pt}
    \setlength{\belowdisplayskip}{3pt}
    C^{\mathcal{G},k}_{t}(i)=C^{\mathcal{G},k}_{t-1}+\log(\hat{P}^{\mathcal{G},k}_{t}(i))-\gamma(i)
\end{equation}
where $i \in \mathcal{G}$ and $k \in [1,K]$. $\gamma(i)$ is the diversity rate. In total, we need to calculate $|V| \times K$ such probabilities for all nodes and beams, where $|V|$ is the number of nodes. Finally, we select top $K$ of them as the predictions. For the location encoder stream, we apply the offset values to the predicted nodes to obtain the precise coordinates.

\section{Experiments}

\subsection{Evaluation Metrics}
\noindent \textbf{Single Future Evaluation.} Same as previous studies \cite{gupta2018social,liang2019peeking,alahi2016social}, we use the following two common evaluation metrics: 1) Average Displacement Error (ADE): the average $L2$ distance between the ground truth locations and predicted locations over all time steps. 2) Final Displacement Error (FDE): the $L2$ distance between the ground truth locations and predicted locations at the final time step.

\noindent \textbf{Multi-future Evaluation.}
We assume that for each data sample, there are $J$ ground truth trajectories and the model makes $K$ predictions. Following the recent public benchmark \cite{liang2020garden} on the Forking Paths dataset, we use: 1) Minimum Average Displacement Error Given $K$ Predictions (minADE$_{K}$); and 2) Minimum Final Displacement Error Given $K$ Predictions (minFDE$_{K}$). For each ground truth $j \in J$ in the data sample $i \in N$, we choose one of the $K$ predictions with the smallest overall distance to $j$ to calculate the average displacement and the one with the smallest final distance to calculate the final displacement.

\noindent \textbf{Percentage of Trajectory Usage.} We propose a new method of evaluation named ``Percentage of Trajectory Usage'' (PTU) to evaluate the comprehensiveness of the performance of multi-future prediction.

\begin{figure}[htp]

\centering

\includegraphics[scale=0.4]{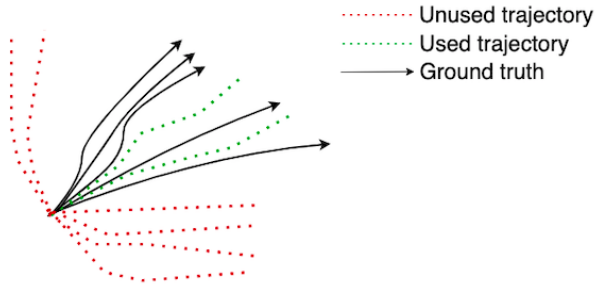}
\setlength{\abovecaptionskip}{-0 cm}
\setlength{\belowcaptionskip}{-0.3 cm}
\caption{An illustration of low percentage of predicted trajectories usage by minADE$_{K}$ evaluation. There are 5 ground truth trajectories (black arrows) and 8 predictions (dotted lines). But only 2 predictions (green dotted lines) are used while evaluating with minADE$_{K}$ and others (red dotted lines) are unused.}

\label{fig:traj_usage}

\end{figure}

Although minADE$_{K}$ and minFDE$_{K}$ can evaluate the displacement between prediction and ground truth, they ignore the diversity of prediction distribution. As illustrated in Fig.~\ref{fig:traj_usage}, there are 5 ground truth trajectories and 8 predicted trajectories. However, according to the definitions of minADE$_{K}$ and minFDE$_{K}$, only 2 predictions are included in the evaluation, which are the closest prediction for each ground truth. We consider this an incomprehensive prediction, where several ground truth trajectories share the same predicted trajectory. Ideally, we expect to have 5 distinct predictions corresponding to 5 ground truth trajectories. In addition, Yuan \etal \cite{yuan2019diverse} developed \textit{Average Self Distance} (ASD) and \textit{Final Self Distance} (FSD) to evaluate the diversity of prediction distribution by calculating the $L2$ distance between each prediction with its nearest prediction. However, diversity of prediction that calculated by ASD and FSD fails to consider the number of predicted trajectories that lie in the distribution of ground truth.

To evaluate the comprehensiveness of prediction distribution, we define PTU as:
\begin{equation}
    \setlength{\abovedisplayskip}{3pt}
    \setlength{\belowdisplayskip}{3pt}
    {\rm PTU}=\frac{\sum^{N}_{i=1} |\hat{p}_{i}| / |Y_{i}| }{N}
    \label{ptu}
\end{equation}
where $|\hat{p}_{i}|$ denotes the number of predictions used while evaluating with minADE$_{K}$ and minFDE$_{K}$ and $\left |Y_{i} \right |$ denotes the number of ground truth trajectories in a data sample. We sum such percentage for all $N$ data samples then average it. Under the same result of minADE$_{K}$ and minFDE$_{K}$, a larger PTU represents a more comprehensive prediction.

\begin{table*}
\begin{center}
\begin{tabular}{ l | c | c | c | c | c | c | c | c }  
 \hline
 \multirow{2}*{Method} &
 \multicolumn{4}{c|}{$\rm minADE_{20}\downarrow$} & 
 \multicolumn{4}{c}{$\rm minFDE_{20}\downarrow$} \\

 \cline{2-9}
 & 45-degree & top-down & all & PTU $\uparrow$ & 45-degree & top-down & all & PTU $\uparrow$ \\ [0.5ex] 
 \hline\hline
 LSTM & 201.0 & 183.7 & 196.7 & N/A & 381.5 & 355.0 & 374.9 & N/A \\
 \hline
 Social-GAN(PV) & 191.2 & 176.5 & 187.5 & 44.70\% & 351.9 & 335.0 & 347.7 & 42.82\% \\
 \hline
 Social-GAN(V) & 187.1 & 172.7 & 183.5 & 43.00\% & 342.1 & 326.7 & 338.3 & 41.85\% \\
 \hline
 Next & 186.6 & 166.9 & 181.7 & N/A & 360.0 & 326.6 & 351.7 & N/A \\
 \hline
 Multiverse & 168.9 & 157.7 & 166.1 & 47.45\% & 333.8 & 316.5 & 329.5 & 44.35\%\\
 \hline
 \textbf{Ours} & \textbf{165.5} & \textbf{154.5} & \textbf{162.8} & \textbf{48.65\%} & \textbf{318.9} & \textbf{302.5} & \textbf{314.8} & \textbf{50.83\%} \\
 \hline

\end{tabular}
\end{center}
\setlength{\abovecaptionskip}{-0.2 cm}
\setlength{\belowcaptionskip}{-0.5 cm}
\caption{Quantitative evaluation of multi-future trajectory prediction. minADE$_{K}$ and minFDE$_{K}$ results are presented on 45-degree, top-down and all views. PTU results are only evaluated for multi-future prediction models. All models are trained on the VIRAT/ActEV dataset and tested on the Forking Paths dataset.}
\label{quantmulti}
\end{table*}

\subsection{Implementation Details}
We use the same data processing method as \cite{gupta2018social} and, following \cite{liang2020garden}, we apply the pre-trained scene segmentation model \cite{chen2017deeplab} to obtain scene segmentation features. We utilize a single convolutional LSTM layer for both the encoder and decoder; and aggregate features of one-hop neighbors in our graph-based transformer. We set the learning rate as 0.3 with decay value of 0.95 and weight decay of 0.001, which are the same as \cite{liang2020garden}. We embed the graph features with an embedding size of 32 and the hidden state sizes for both the encoder and decoder are 256. For hyperparameters $\alpha$ and $\beta$ in total loss, we set $\alpha=1.0$ and $\beta=0.2$; and $\mu$ in our exponential smooth $L1$ loss is 10. We only apply such exponential loss on single-future prediction as it will affect the diversity of multi-future prediction through experiments. In order to align with \cite{liang2020garden}, we also generate $K=20$ most possible predictions for each data sample in multi-future prediction.

\subsection{Multi-future Prediction}
\noindent \textbf{The Forking Paths Dataset.} The Forking Paths dataset \cite{liang2020garden} is a simulated dataset specifically designed for multi-future prediction. This dataset was constructed by 5 scenes in VIRAT/ActEV and 4 scenes in ETH/UCY. There are 127 scenarios, each of which is rendered in three 45-degree views and one top-down view. There is one controlled agent in each scenario which has on average 5.9 future trajectories. We aim to predict multiple trajectories for each controlled agent. The length of observation time is $T_{obs}=8$ frames and prediction time is $T_{pred-obs}=12$ frames.

\noindent \textbf{Baselines.} We compare our model with 4 baseline models. \textbf{\textit{LSTM}}: A simple LSTM implementation which only models the trajectory inputs. \textbf{\textit{Social GAN}} \cite{gupta2018social}: A recent GAN-based model that generates multimodal prediction distributions. We report two configurations: the model with only variety loss (Social-GAN(V)) and with both variety loss and global pooling (Social-GAN(PV)). \textbf{\textit{Next}} \cite{liang2019peeking}: the state-of-the-art model on the VIRAT/ActEV for single-future prediction. Since the model utilizes rich visual features, we compare our model with Next without activity prediction module. \textbf{\textit{Multiverse}} \cite{liang2020garden}: the recent state-of-the-art probabilistic model for multi-future prediction on the Forking Paths dataset.

\noindent \textbf{Quantitative Evaluation.} Table \ref{quantmulti} shows the comparisons of multi-future prediction between basline models and our model in minADE$_{20}$, minFDE$_{20}$ and PTU metrics. We can see that our model outperforms all baseline methods. Compared to the current state-of-the-art model Multiverse, the average minADE$_{20}$ reduces by 3 points and the average minFDE$_{20}$ reduces by 15 points on all views. The PTU value is higher than Multiverse under minADE$_{K}$ by 1.2\% and under minFDE$_{K}$ by 6.5\%, which demonstrates that our model generates more comprehensive predictions.

\begin{figure*}[htbp]
    \centering
    \subfloat[Imperfect/Error cases of Multiverse]{
    \label{multvierse}
    \includegraphics[width=2.5cm,height=2cm]{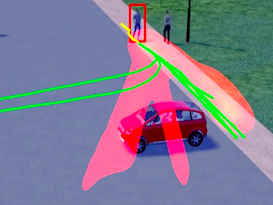}
    \includegraphics[width=2.5cm,height=2cm]{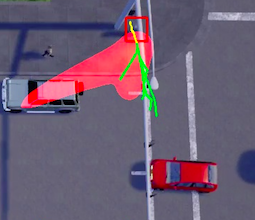}
    \includegraphics[width=2.5cm,height=2cm]{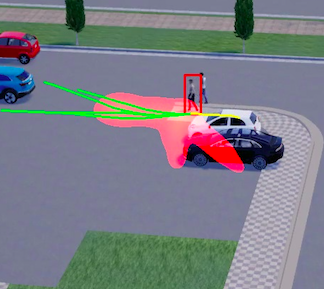}
    \includegraphics[width=2.5cm,height=2cm]{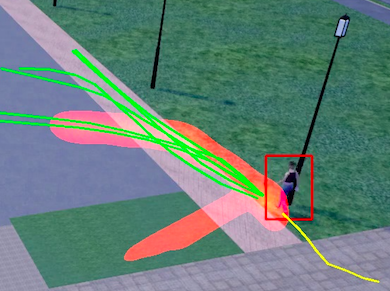}
    \includegraphics[width=2.5cm,height=2cm]{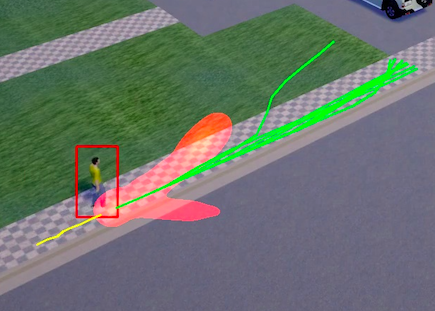}
    \includegraphics[width=2.5cm,height=2cm]{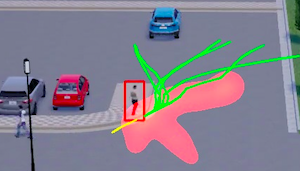}
    }
    
    \subfloat[Improved cases of our model]{
    \label{ours}
    \includegraphics[width=2.5cm,height=2cm]{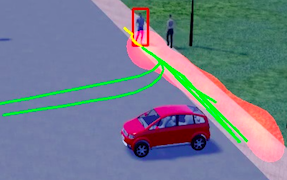}
    \includegraphics[width=2.5cm,height=2cm]{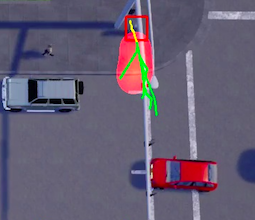}
    \includegraphics[width=2.5cm,height=2cm]{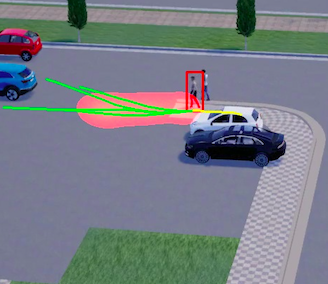}
    \includegraphics[width=2.5cm,height=2cm]{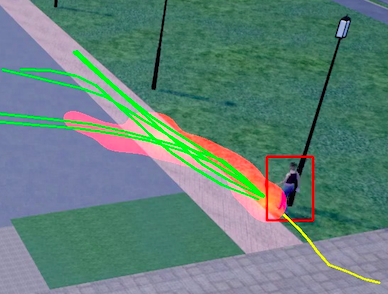}
    \includegraphics[width=2.5cm,height=2cm]{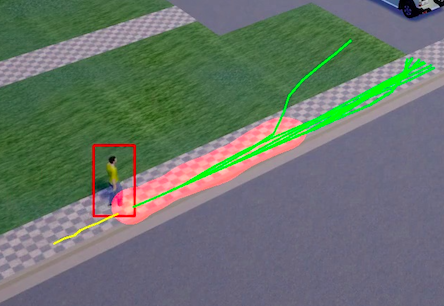}
    \includegraphics[width=2.5cm,height=2cm]{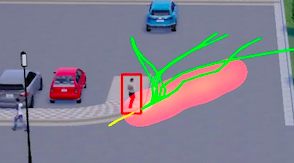}
    }

    \setlength{\abovecaptionskip}{0.cm}
    \setlength{\belowcaptionskip}{-0.3 cm}
    \caption{Qualitative comparisons between the Multiverse and our model. The yellow and green lines are observed and ground truth trajectories. The heatmaps are prediction distributions.}
    \label{fig:qualitative}
\end{figure*}

\noindent \textbf{Qualitative Evaluation.}
Fig. \ref{fig:qualitative}\subref{multvierse} are the results of Multiverse and Fig. \ref{fig:qualitative}\subref{ours} are those of ours. From the three sets of comparisons on the left, we can see that the predictions of Multiverse go through the vehicles, whereas ours can make predictions that lie in the ground truth distribution without colliding with other objects. These cases demonstrate that our spatial transformer can detect objects and make reasonable decisions accordingly. Furthermore, the three sets on the right in Fig. \ref{fig:qualitative}\subref{multvierse} show the temporal inconsistent cases of Multiverse, whereas ours can make smooth predictions, which reflects that our Memory Replay is effective in retaining the temporal consistency of predictions.

\subsection{Single Future Prediction}
\noindent \textbf{VIRAT/ActEV Dataset.}
Following \cite{liang2020garden}, we use VIRAT/ActEV \cite{awad2018trecvid} as the dataset for single-future trajectory prediction. This dataset is designed to evaluate tasks such as activity detection and object tracking. We use the same training, validation and testing split as \cite{liang2019peeking,liang2020garden,liang2020simaug} to make fair comparisons. Observation length is 3.2 seconds (8 frames) and prediction length is 4.8 seconds (12 frames), which are the same as the previous works \cite{alahi2016social,gupta2018social,liang2019peeking,liang2020simaug,liang2020garden}.

\begin{table}
\begin{center}
\begin{tabular}{c || c | c } 
 \hline
 Method & ADE$\downarrow$ & FDE$\downarrow$ \\ [0.5ex] 
 \hline\hline
 LSTM & 23.98 & 44.97 \\ 
 \hline
 Social-GAN(V) & 30.40 & 61.93 \\ 
 \hline
 Social-GAN(PV) & 30.42 & 60.70 \\ 
 \hline
 Next & 19.78 & 42.43 \\ 
 \hline
 Multiverse & \textbf{18.51} & \textbf{35.84} \\ 
 \hline
 Ours & 18.58 & 36.08 \\ 
 \hline
 
\end{tabular}
\end{center}
\setlength{\abovecaptionskip}{-0.2 cm}
\setlength{\belowcaptionskip}{0.7 cm}
\caption{Quantitative evaluation of single-future trajectory prediction on VIRAT/ActEV dataset in ADE and FDE metrics.}
\label{quantsingle}
\end{table}

\begin{table}
\begin{center}
\begin{tabular}{c || c c} 
 \hline
 Method & ADE$\downarrow$ & FDE$\downarrow$ \\ [0.5ex] 
 \hline\hline
 Single-scale graph & 19.71 & 37.32 \\ 
 \hline
 No Location Encoder & 41.18 & 61.23 \\
 \hline
 No Memory Replay & 19.34 & 37.05 \\
 \hline
 No exponential loss & 19.39 & 37.09 \\
 \hline
 \textbf{Full Model} & \textbf{18.58} & \textbf{36.08}\\
 \hline
\end{tabular}
\end{center}
\setlength{\abovecaptionskip}{-0.2 cm}
\setlength{\belowcaptionskip}{0.1 cm}
\caption{Ablation study for key components in our model on single-future prediction.}
\label{abl_single}
\end{table}

\begin{table*}
\begin{center}
\begin{tabular}{l | c | c | c | c | c | c | c | c }  
 \hline
 \multirow{2}*{Method} &
 \multicolumn{4}{c|}{$\rm minADE_{20}\downarrow$} & 
 \multicolumn{4}{c}{$\rm minFDE_{20}\downarrow$} \\

 \cline{2-9}
 & 45-degree & top-down & all & PTU $\uparrow$ & 45-degree & top-down & all & PTU $\uparrow$ \\ [0.5ex] 
 \hline\hline
 Single-scale graph & 170.9 & 160.3 & 168.3 & 44.89\% & 337.9 & 315.8 & 332.4 & 39.96\% \\
 No Location Encoder & 245.4 & 237.3 & 243.4 & 31.14\% & 463.0 & 441.7 & 457.7 & 28.56\% \\
 No Memory Replay & 167.3 & \textbf{153.2} & 163.7 & 46.53\% & 330.1 & 306.2 & 324.1 & 42.43\% \\
 \hline
 \textbf{Ours (Full Model)} & \textbf{165.5} & 154.5 & \textbf{162.8} & \textbf{48.65\%} & \textbf{318.9} & \textbf{302.5} & \textbf{314.8} & \textbf{50.83\%} \\
 \hline

\end{tabular}
\end{center}
\setlength{\abovecaptionskip}{-0.2 cm}
\setlength{\belowcaptionskip}{-0.5 cm}
\caption{Ablation study for key components in our model on multi-future prediction.}
\label{abl_multi}
\end{table*}

\noindent \textbf{Quantitative Evaluation.}
The results of single-future prediction are shown in Table \ref{quantsingle}. The result of our model is the second best and very close to that of Multiverse, exhibiting large improvement over GAN-based models. This implies that our model performs effectively for both simulated multi-future and real-world single-future scenarios.

\subsection{Ablation Study}
\noindent \textbf{Ablations of key components.} For single-future prediction, we verify four components: without multi-scale graph and only keep a scale of 36 $\times$ 18, no Memory Replay module, no Location Encoder, and no Exponential Smooth $L1$ loss. For multi-future prediction, we only test on the first three components as the exponential loss is only designed for promoting performance of single-future prediction in training. As indicated in Tables \ref{abl_single} and \ref{abl_multi}, without any of these key components, our model shows performance drop in varying degrees. Additionally, we apply PTU on all key components on multi-future prediction. From Table \ref{abl_multi}, we can see that our model achieves the highest PTU under both minADE$_{K}$ and minFDE$_{K}$.

\begin{table}
\begin{center}
\begin{tabular}{c || c c} 
 \hline
 Value of $\mu$ & ADE$\downarrow$ & FDE$\downarrow$ \\ [0.5ex] 
 \hline\hline
 $+\infty$ & 19.39 & 37.09 \\ 
 \hline
 $\mu$=20 & 19.19 & 36.83 \\ 
 \hline
 $\mu$=10 & 18.58 & \textbf{36.08} \\ 
 \hline
 $\mu$=5 & \textbf{18.56} & 36.24 \\ 
 \hline
\end{tabular}
\end{center}
\setlength{\abovecaptionskip}{-0.2 cm}
\setlength{\belowcaptionskip}{0.1 cm}
\caption{Ablation study for different values of $\mu$ in our Exponential Smooth $L1$ loss on single-future prediction.}
\label{abl_gamma}
\end{table}

\noindent \textbf{Exponential smooth $L1$ loss.} Multiplying the penalty term can lead to small improvement, as it guides the model to focus more on the earlier data in a sequence, which can affect the overall performance of predictions in the whole sequence.
We select the values of $\mu$ as $+\infty$, 20, 10 and 5 ($+\infty$ means only using smooth $L1$ loss) for comparisons. Table \ref{abl_gamma} shows that when $\mu=10$, our model achieves the overall best performance, which reduces the ADE by $3.6\%$ and FDE by $2.4\%$ on average.

\begin{figure}[htbp]
    \centering
    \subfloat[]{
    \label{misdirect}
    \includegraphics[width=3.3cm,height=2.3cm]{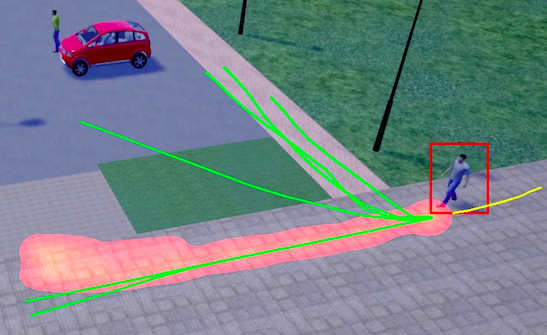}
    }
    \quad
    \subfloat[]{
    \label{speed}
    \includegraphics[width=3.3cm,height=2.3cm]{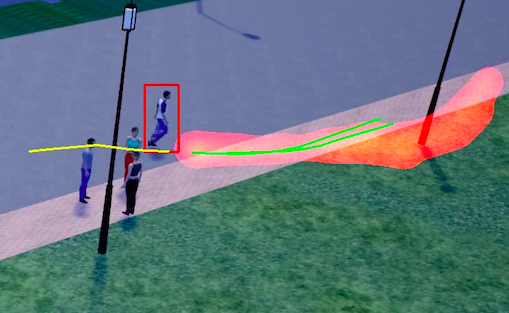}
    }
    
    \setlength{\abovecaptionskip}{0. cm}
    \setlength{\belowcaptionskip}{-0.5 cm}
    \caption{Example of limitations of our model.}
    \label{fig:limits}
\end{figure}

\noindent \textbf{Limitations.} We show some imperfect cases of our model as limitations. Fig. \ref{fig:limits}\subref{misdirect} is a case that our model only predicts trajectories that go straight ahead, whereas there are some ground truth trajectories turning right in diverse degrees. It might be improved if diversity control is applied on loss functions or during selecting final predictions. Fig. \ref{fig:limits}\subref{speed} shows our model sometimes is not aware of the walking speed and makes prediction apparently longer than the ground truth. Data augmentation would help reduce similar occurrences. These imperfect cases and possible ideas would inspire our future research.
\vspace{-0.1cm}

\section{Conclusion}
This paper focuses on multi-future pedestrian trajectory prediction when there are multiple plausible future trajectories for each pedestrian in the ground truth. We model the spatial interactions by a graph-based spatial transformer, which utilizes an improved attention-based message generation and aggregation method as well as adopting a multi-scale graph structure. We also introduce the Memory Replay algorithm to generate smooth trajectories by coordinating with the transformer. Moreover, we propose Percentage of Trajectory Usage to evaluate the comprehensiveness of multi-future prediction. Our proposed model achieves state-of-the-art performance for multi-future prediction on the Forking Paths dataset and our single-future prediction results can compete with those by current state-of-the-art models on the VIRAT/ActEV dataset.

\clearpage

{\small
\bibliographystyle{ieee_fullname}
\bibliography{egbib}
}

\end{document}